\documentclass[11pt,a4paper]{article}
\usepackage[hyperref]{acl2018}
\usepackage{times}
\usepackage{latexsym}
\usepackage{graphicx}
\usepackage{subfigure}
\usepackage{url}
\usepackage{color}

\aclfinalcopy 

\title{A Web-scale system for scientific knowledge exploration}

\author{Zhihong Shen \\
  Microsoft Research  \\
  Redmond, WA, USA \\
  \And
  Hao Ma \\
  Microsoft Research  \\
  Redmond, WA, USA \\
  {\tt \{zhihosh,haoma,kuansanw\}@microsoft.com} \\ \And
  Kuansan Wang \\
  Microsoft Research  \\
  Redmond, WA, USA \\
  }
 
\date{}

\begin{document}
\maketitle
\begin{abstract}
 
To enable efficient exploration of Web-scale scientific knowledge, it is necessary to organize scientific publications into a hierarchical concept structure. 
In this work, we present a large-scale system to (1) 
identify hundreds of thousands of scientific concepts, (2) tag these identified concepts to hundreds of millions of scientific publications by leveraging both text and graph structure,   and (3) build a six-level concept hierarchy with a subsumption-based model. 
The system builds the most comprehensive cross-domain scientific concept ontology published to date, with more than 200 thousand concepts and over one million relationships. 

\end{abstract}

\section{Introduction}

Scientific literature has grown exponentially over the past centuries, with a two-fold increase every 12 years~\cite{dong2017century}, and millions of new publications are added every month.
Efficiently identifying relevant research has become an ever increasing challenge due to the unprecedented growth of scientific knowledge.  In order to assist researchers to navigate the entirety of scientific information, we present a deployed system that organizes scientific knowledge in a hierarchical manner.

To enable a streamlined and satisfactory semantic exploration experience of scientific knowledge, three criteria must be met: 

\begin{itemize}
\item a comprehensive coverage on the broad spectrum of academic disciplines and concepts (we call them \emph{concepts} or \emph{fields-of-study}, abbreviated as \emph{FoS}, in this paper);
\item a well-organized hierarchical structure of scientific \emph{concepts};
\item an accurate mapping between these \emph{concepts} and all forms of academic \emph{publications}, including books, journal articles, conference papers, pre-prints, etc.
\end{itemize}

\begin{figure}[!t]
\centerline{\includegraphics[width=8cm]{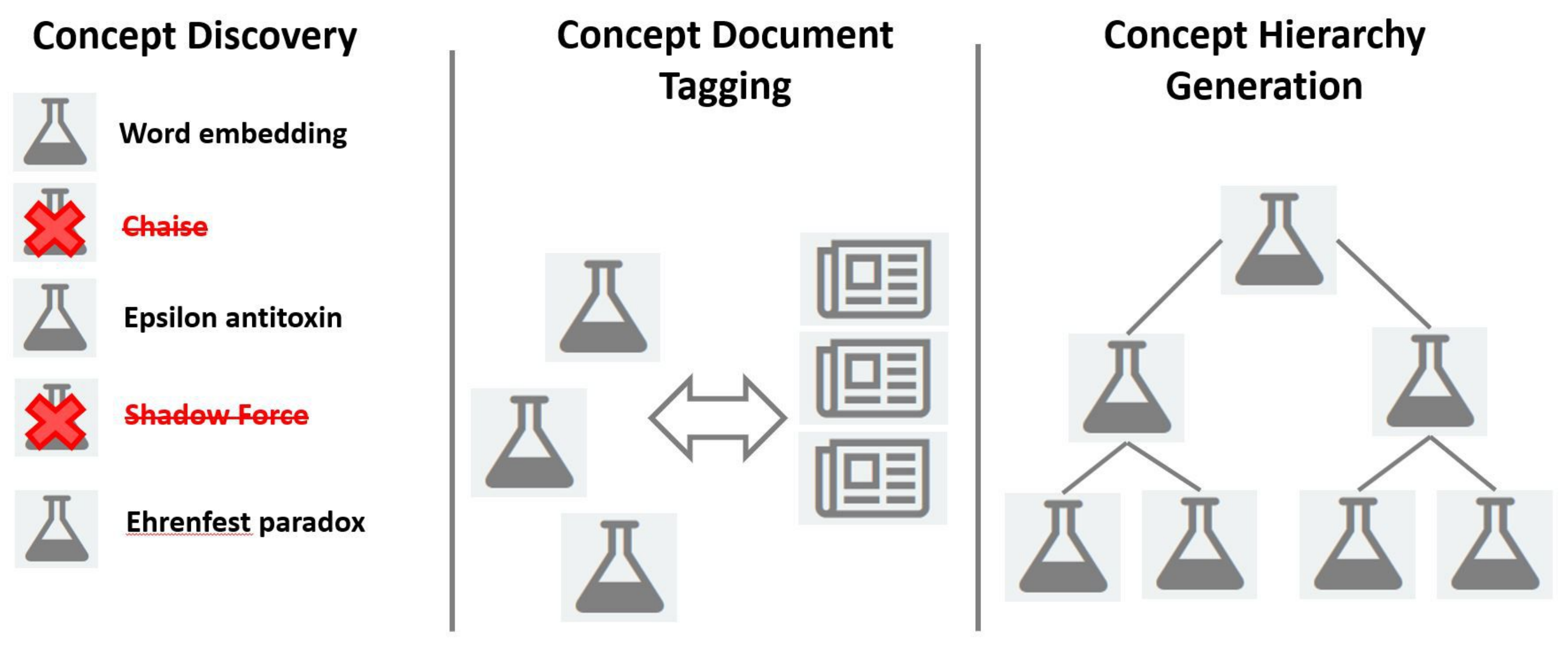}}
\caption{Three modules of the system: \emph{concept discovery}, \emph{concept-document tagging}, and \emph{concept hierarchy generation}.}
    \label{fig:FoSTagging_artchitecture}
\end{figure}

\begin{table*}[!t]
\centering
\begin{tabular}{|c|c|c|c|}
\hline
& {\bf Concept } & {\bf Concept } & {\bf Hierarchy }\\
& {\bf  discovery} & {\bf tagging} & {\bf building}\\\hline
{\bf Main } & scalability / trustworthy  & scalability / & stability / \\
{\bf challenges} &   representation &  coverage & accuracy \\ \hline
{\bf Problem } & knowledge base &  multi-label  & topic hierarchy \\
{\bf formulation} & type prediction &  text classification & construction \\ \hline
{\bf Solution / } & Wikipedia / KB /  & word embedding /  & extended \\ 
{\bf model(s)} & graph link analysis & text + graph structure  & subsumption \\ \hline
{\bf Data scale} & $10^5$ -- $10^6$ & $10^9$ -- $10^{10}$ & $10^6$ -- $10^7$ \\ \hline
{\bf Data update} &  &  &  \\
{\bf frequency} & monthly & weekly & monthly \\ \hline
\end{tabular} 

\caption{System key features at a glance.}\vspace{-0.1in}\label{tab:System_summary}
\end{table*}

To build such a system on Web-scale, the following challenges need to be tackled:

\begin{itemize}
\item \textbf{Scalability:}  Traditionally, academic discipline and concept taxonomies have been curated manually on a scale of hundreds or thousands, 
 which is insufficient in modeling the richness of academic concepts across all domains. Consequently, the low concept coverage also limits the exploration experience of hundreds of millions of scientific publications. 
 
\item \textbf{Trustworthy representation:} Traditional concept hierarchy construction approaches extract concepts from unstructured documents, select representative terms to denote a concept, and  build the hierarchy on top of them~\cite{sanderson1999deriving, liu2012automatic}. 
The concepts extracted this way not only lack authoritative definition, but also contain erroneous topics with subpar quality which is not suitable for a production system.

\item \textbf{Temporal dynamics:} Academic publications are growing at an unprecedented pace (about 70K more papers per day according to our system) and new concepts are emerging faster than ever. 
This requires frequent inclusion on latest publications and re-evaluation in tagging and hierarchy-building results. 
\end{itemize}

In this work, we present a Web-scale system with three modules---concept discovery, concept-document tagging, and concept-hierarchy generation---to facilitate scientific knowledge exploration (see Figure~\ref{fig:FoSTagging_artchitecture}). 
This is one of the core components in constructing the Microsoft Academic Graph (MAG), which enables a semantic search experience in the 
academic domain\footnote{The details about where and how we obtain, aggregate, and ingest academic publication information into the system is out-of-scope for this paper and for more information please refer to \cite{sinha2015overview}.}. 
MAG is a scientific knowledge base and a heterogeneous graph with six types of academic entities: publication, author, institution, journal, conference, and field-of-study (i.e., \emph{concept} or \emph{FoS}). As of March 2018, it contains more than 170 million publications with over one billion paper citation relationships, and is the largest publicly available academic dataset to date\footnote{\url{https://www.openacademic.ai/oag/}}. 

To generate high-quality \emph{concepts} with comprehensive coverage, we leverage Wikipedia articles as the source of concept discovery. Each Wikipedia article is an \emph{entity} in a general knowledge base (\emph{KB}). A \emph{KB} entity associated with a Wikipedia article is referred to as a Wikipedia entity. We formulate concept discovery as a knowledge base type prediction problem~\cite{neelakantan2015inferring} 
and use graph link analysis to guide the process. In total, 228K academic concepts are identified 
from over five million English Wikipedia entities. 

During the tagging stage, both textual information and graph structure are considered. The text from Wikipedia articles and papers' meta information (e.g., titles, keywords, and abstracts) are used as the \emph{concept}'s and \emph{publication}'s textual representations respectively. 
Graph structural information is leveraged by using text from a \emph{publication}'s neighboring nodes in MAG (its citations, references, and  publishing venue) as part of the \emph{publication}'s representation with a discounting factor.  
We limit the search space for each publication to a constant range, reduce the complexity to $O(N)$ for scalability, where $N$ is the number of publications.  Close to one billion concept-publication pairs are established with associated confidence scores.

Together with the notion of subsumption~\cite{sanderson1999deriving}, this confidence score is then used to construct a six-level directed acyclic graph (DAG) 
hierarchy with over 200K nodes and more than one million edges.

Our system is a deployed product with regular data refreshment and algorithm improvement. Key features of the system are summarized in Table~\ref{tab:System_summary}. 
The system is updated weekly or monthly to include fresh content on the Web. Various document and language understanding techniques are experimented with and incorporated to incrementally improve the performance over time.  

\section{System Description}
\subsection{Concept Discovery}
As top level disciplines are extremely important and highly visible to system end users, we manually define 19 top level (``L0") disciplines (such as \emph{physics}, \emph{medicine}) and 294 second level (``L1") sub-domains (examples are \emph{machine learning}, \emph{algebra}) by referencing existing classification\footnote{\url{http://science-metrix.com/en/classification}} and get their correspondent Wikipedia entities in a general in-house knowledge base (\emph{KB}). 

It is well understood that entity types in a general \emph{KB} are limited and far from complete. 
Entities labeled with \emph{FoS} type in \emph{KB} are in the lower thousands and noisy for both in-house \emph{KB} and latest Freebase dump\footnote{\url{https://developers.google.com/freebase/}}. The goal is to identify more \emph{FoS} type entities from over 5 million English Wikipedia entities in an in-house KB. We formulate this task as a knowledge base type prediction problem, and focus on predicting only one specific type---\emph{FoS}.

In addition to the above-mentioned ``L0" and ``L1" FoS,
we manually review and identify over 2000 high-quality ones as initial seed FoS. 
We iterate a few rounds between a \emph{graph link analysis} step for candidate exploration and an \emph{entity type based filtering and enrichment} step for candidate fine-tuning based on \emph{KB} types. 

\emph{\textbf{Graph link analysis}}:
To drive the process of exploring new FoS candidates, we apply the intuition that if the majority of an entity's nearest neighbours are FoS, then it is highly likely an FoS as well. 
To calculate nearest neighbours, a distance measure between two Wikipedia entities is required. We use an effective and low-cost approach based on Wikipedia link analysis to compute the semantic closeness~\cite{milne2008effective}. 
We label a Wikipedia entity as an FoS candidate if there are more than $K$ neighbours in its top $N$ nearest ones are in a current FoS set. Empirically, $N$ is set to 100 and $K$ is in [35, 45] range for best results. 

\emph{\textbf{Entity type based filtering and enrichment}}: 
The candidate set generated in the above step contains various types of entities, such as \emph{person}, \emph{event}, \emph{protein}, \emph{book topic}, etc.\footnote{Entity types are obtained from the in-house KB, which has higher type coverage compared with Freebase, details on how the in-house KB produces entity types is out-of-scope and not discussed in this paper.}
Entities with obvious invalid types are eliminated (e.g. \emph{person}) and entities with good types are further included (e.g. \emph{protein}, such that all Wikipedia entities which have labeled type as \emph{protein} are added). The results of this step are used as the input for \emph{graph link analysis} in the next iteration. 

More than 228K FoS have been identified with this iterative approach, based on over 2000 initial seed FoS.

\subsection{Tagging Concepts to Publications}

We formulate the concept tagging as a multi-label classification problem; i.e. each publication could be tagged with multiple FoS as appropriate. In a naive approach, the complexity could reach $M\cdot N$ to exhaust all possible pairs, where $M$ is 200K+ for FoS and $N$ is close to 200M for publications. Such a naive solution is computationally expensive and wasteful, since most scientific publications cover no more than 20 FoS based on empirical observation. 

We apply heuristics to cut candidate pairs aggressively to address the scalability challenge, to a level of 300--400 FoS per publication\footnote{We include all L0s and L1s and FoS entities spotted in a publication's \emph{\textbf{extended} representing text}, which is defined later in this section}. Graph structural information is incorporated in addition to textual information to improve the accuracy and coverage when limited or inadequate text of a \emph{concept} or \emph{publication} is accessible. 

We first define \emph{\textbf{simple} representing text} (or \emph{SRT}) and \emph{\textbf{extended} representing text} (or \emph{ERT}). 
\emph{SRT} is the text used to describe the academic entity itself. \emph{ERT} is the extension of \emph{SRT} and leverages the graph structural information to include textual information from its neighbouring nodes in MAG.

A publishing venue's full name (i.e. the journal name or the conference name) is its \emph{SRT}.
The first paragraph of a concept's Wikipedia article is used as its \emph{SRT}. Textual meta data, such as title, keywords, and abstract is a publication's \emph{SRT}. 

 We sample a subset of publications from a given venue and concatenate their \emph{SRT}. This is used as this venue's \emph{ERT}.  For broad disciplines or domains (e.g. L0 and L1 FoS), Wikipedia text becomes too vague and general to best represent its academic meanings. We manually curate such concept-venue pairs and aggregate \emph{ERT} of venues associated with a given concept to obtain the \emph{ERT} for the concept. 
 For example, \emph{SRT} of a subset of papers from \emph{ACL} are used to construct \emph{ERT} for \emph{ACL}, and subsequently be part of the \emph{ERT} for \emph{natural language processing} concept.
A \emph{publication}'s \emph{ERT} includes \emph{SRT} from its citations, references and \emph{ERT} of its linked publishing \emph{venue}. 

We use $h_{s}^{p}$ and $h_{e}^{p}$ to denote the representation of a \emph{publication} ($p$)'s \emph{SRT} and \emph{ERT},  $h_{s}^{v}$ and $h_{e}^{v}$ for a \emph{venue} ($v$)'s \emph{SRT} and \emph{ERT}. Weight $w$ is used to discount different neighbours' impact as appropriate. Equation \ref{eq:Pub_aggre} and \ref{eq:Venue_aggre} formally define publication \emph{ERT} and venue \emph{ERT} calculation. 

\begin{equation} \label{eq:Pub_aggre}
h_{e}^{p} = h_{s}^{p} + \sum_{i \in Cit}w_{i}h_{s}^{p}(i) +
\sum_{j \in Ref}w_{j}h_{s}^{p}(j) + w_{v}h_{e}^{v}
\end{equation}

\begin{equation} \label{eq:Venue_aggre}
h_{e}^{v} =\sum_{i \in V}h_{s}^{p}(i) + h_{s}^{v}
\end{equation}

Four types of features are extracted from the text: 
bag-of-words (BoW), bag-of-entities (BoE), embedding-of-words (EoW), and embedding-of-entities (EoE).
These features are concatenated for the vector representation $h$ used in Equation \ref{eq:Pub_aggre} and \ref{eq:Venue_aggre}.  
The \emph{confidence score} of a concept-publication pair is the cosine similarity between these vector representations. 

We pre-train the word embeddings by using the skip-gram ~\cite{mikolov2013distributed} on the academic corpus, with 13B  words based on 130M titles and 80M abstracts from English scientific publications. The resulting model contains 250-dimensional vectors for 2 million words and phrases. We compare our model with pre-trained embeddings based on general text (such as Google News\footnote{\url{https://code.google.com/archive/p/word2vec/}} and Wikipedia\footnote{\url{https://fasttext.cc/docs/en/pretrained-vectors.html}}) and observe that the model trained from academic corpus performs better with higher accuracy on the concept-tagging task with more than 10\% margin. 

Conceptually, the calculation of publication and venue's \emph{ERT} is to leverage   neighbours' information to represent itself. The MAG contains hundreds of millions of nodes with billions of edges, hence it is computationally prohibitive by 
optimizing the node latent vector and weights simultaneously. Therefore, in Equation~\ref{eq:Pub_aggre} and~\ref{eq:Venue_aggre}, we initialize $h_{s}^{p}$ and $h_{s}^{v}$ based on textual feature vectors defined above and adopt empirical weight values to directly compute $h_{e}^{p}$ and $h_{e}^{v}$ to make it scalable. 

After calculating the similarity for about 50 billion pairs, 
close to 1 billion are finally picked based on the threshold set by the confidence score. 

\begin{figure}[!t]
\centerline{\includegraphics[width=7cm]{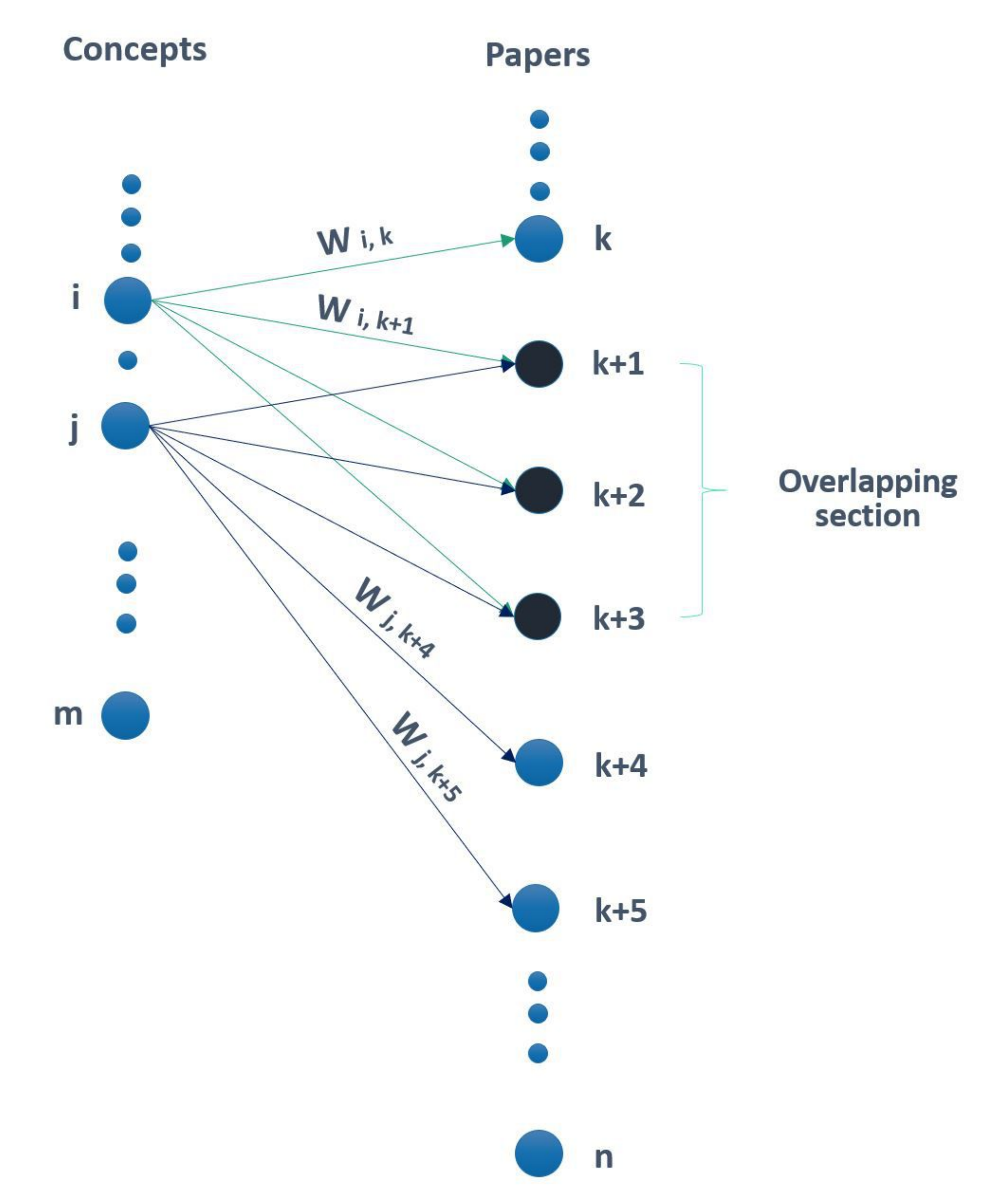}}
\vspace{-0.1in}
\caption{Extended subsumption for hierarchy generation.}\vspace{-0.1in}\label{fig:Hierarchy_explain}
\end{figure}

\subsection{Concept Hierarchy Building}

In this subsection, we describe how to build a concept hierarchy based on concept-document tagging results.
We extend Sanderson and Croft's early work \shortcite{sanderson1999deriving} which uses the notion of subsumption---a form of co-occurrence---to associate related terms. We say term $x$ subsumes $y$ if $y$ occurs only in a subset of the documents that $x$ occurs in. In the hierarchy, $x$ is the parent of $y$. In reality, it is hard for $y$ to be a strict subset of $x$. Sanderson and Croft's work  relaxed the subsumption to 80\% (e.g. $P(x|y) \geq 0.8, P(y|x) < 1$).  

In our work, 
we extend the concept co-occurrence calculation weighted with the concept-document pair's 
confidence score from previous step. More formally, we define a \emph{weighted relative coverage} score between two concepts $i$ and $j$ as below and illustrate in Figure \ref{fig:Hierarchy_explain}. 

\begin{equation}
RC(i,j)=
\frac{\sum_{k \in (I \cap J)} w_{i,k}}{\sum_{k \in I} w_{i,k}} 
- 
\frac{\sum_{k \in (I \cap J)} w_{j,k}}{\sum_{k \in J} w_{j,k}}
\end{equation}

\begin{figure}[!t]
\centerline{\includegraphics[width=8cm]{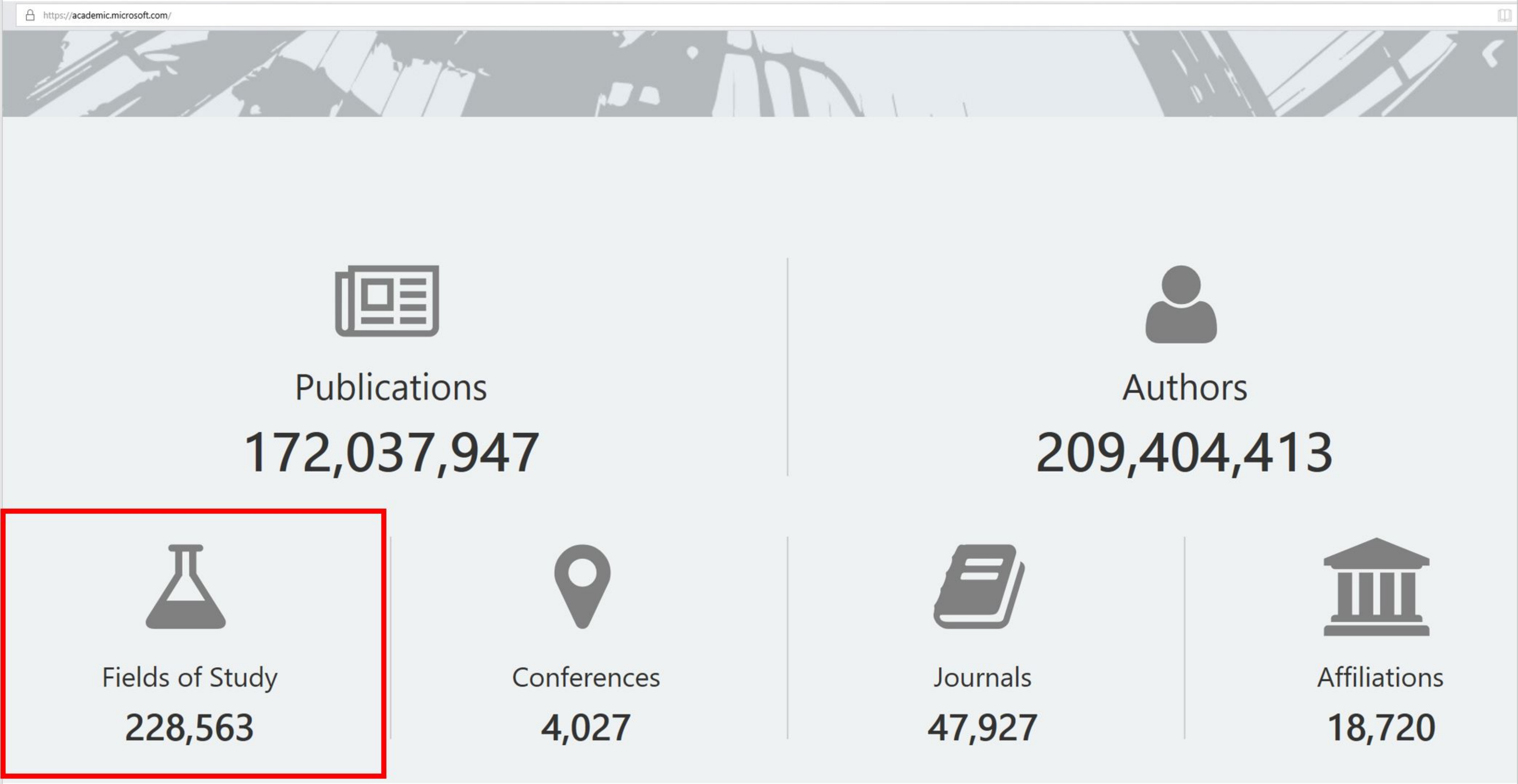}}

\caption{Deployed system homepage at March 2018, with all six types of entities statistics: over 228K \emph{fields-of-study}.}\label{fig:System_Deploy}
\end{figure}

Set $I$ and $J$ are documents tagged with concepts $i$ and $j$ respectively. $I \cap J$ is the overlapping set of documents that are tagged with both $i$ and $j$.  $w_{i,k}$ denotes the confidence score (or weights) between concept $i$ and document $k$, which is the final \emph{confidence score} in the previous concept-publication tagging stage.
When $RC(i,j)$ is greater than a given positive threshold\footnote{It is usually in [$0.2$, $0.5$] based on empirical observation.}, $i$ is the child of $j$.
Since this approach does not enforce single parent for any FoS, it results in a directed acyclic graph (DAG) hierarchy. 

With the proposed model, we construct a six level FoS hierarchy (from L0 and L5) on over 200K concepts with more than 1M parent-child pairs. Due to the high visibility, high impact and small size, the hierarchical relationships between L0 and L1 are manually inspected and adjusted if necessary. The remaining L2 to L5 hierarchical structures are produced completely automatically by the extended subsumption model. 

One limitation of
subsumption-based models is the intransitiveness of parent-child relationships.
This model also lacks a type-consistency check between parents and children.
More discussions on such limitations with examples will be in evaluation section~\ref{sec:Evaluation}.

\section{Deployment and Evaluation}

\begin{figure*}[!t]
\centerline{\includegraphics[width=11cm]{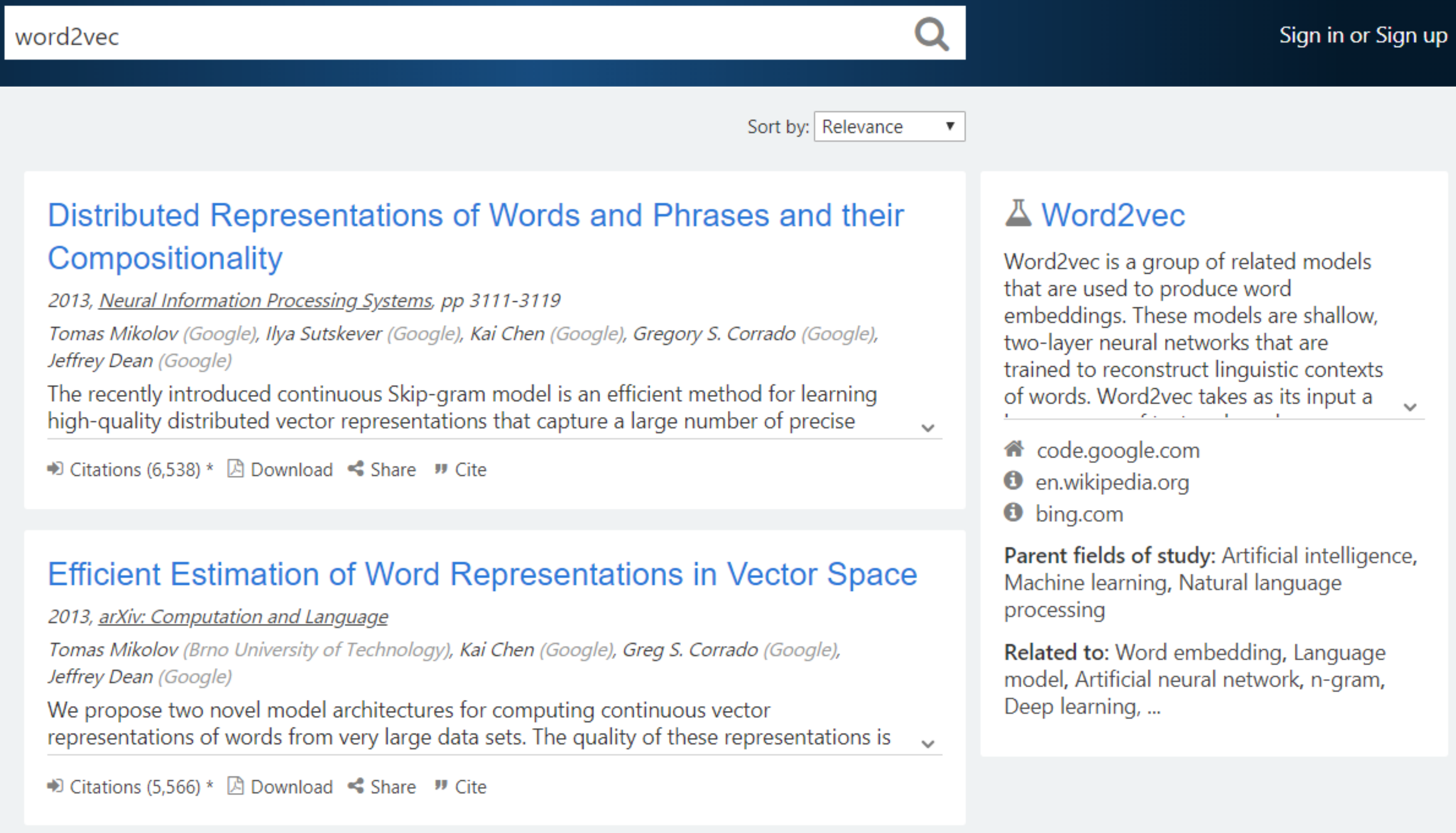}}
\caption{\emph{Word2vec} example, with its parent \emph{FoS},  related \emph{FoS} and top tagged publications.}
    \label{fig:Word2vec_example} 
\end{figure*}

\subsection{Deployment}

The work described in this paper has been deployed in the production system of Microsoft Academic Service\footnote{\url{https://academic.microsoft.com/}}. Figure \ref{fig:System_Deploy} shows the website homepage with entity statistics. 
The contents of MAG, including the full list of FoS, FoS hierarchy structure, and FoS tagging to papers, are accessible via API, website, and full graph dump from \emph{Open Academic Society}\footnote{\url{https://www.openacademic.ai/oag/}}.

Figure \ref{fig:Word2vec_example} shows the example for \emph{word2vec} concept. Concept definition with linked Wikipedia page, its immediate parents (\emph{machine learning}, \emph{artificial intelligence}, \emph{natural language processing}) in the hierarchical structure and its related concepts\footnote{Details about how to generate related entities are out-of-scope and not included in this paper.} (\emph{word embedding}, \emph{artificial neural network}, \emph{deep learning}, etc.) are shown on the right rail pane. Top tagged publications (without \emph{word2vec} explicitly stated in their text) are recognized via graph structure information based on citation relationship.

\begin{table}[!t]
\centering
\small
\begin{tabular}{|l|c|}
\hline
{\bf Step} & {\bf Accuracy}\\\hline
\verb|1. Concept discovery| & {94.75\%} \\
\verb|2. Concept tagging| & {81.20\%} \\
\verb|3. Build hierarchy| & {78.00\%} \\ \hline
\end{tabular}
\caption{Accuracy results for each step.}\label{tab:eval_accuracy_results}
\end{table}

\begin{table*} [!t]
\centering
\small
\begin{tabular}{|c|c|c|c|c|c|}
\hline
{\bf L5} & {\bf L4} & {\bf L3} & {\bf L2} & {\bf L1} & {\bf L0} \\\hline
Convolutional Deep  & Deep belief & Deep  & Artificial  & Machine  & Computer \\
Belief Networks & network &  learning &  neural network &  learning &  Science\\\hline
(Methionine synthase)  & Methionine  &  & Amino  & Biochemistry / & Chemistry / \\
 reductase &  synthase & Methionine &  acid & Molecular biology & Biology \\\hline
 (glycogen-synthase-D)  & Phosphorylase & Glycogen  &  &  &  \\
phosphatase & kinase &  synthase & Glycogen & Biochemistry & Chemistry \\\hline
 & Fr\'{e}chet  & Generalized extreme  & Extreme  &  &  \\
 &  distribution & value distribution &  value theory & Statistics & Mathematics \\\hline
Hermite's  & Hermite  & Spline  &  & Mathematical  & \\
 problem &  spline &  interpolation & Interpolation &  analysis & Mathematics\\\hline
\end{tabular}
\caption{Sample results for \emph{FoS} hierarchy.}\vspace{-0.1in}\label{tab:example_results}
\end{table*}

\subsection{Evaluation}
\label{sec:Evaluation}

For this deployed system, we evaluate the accuracy on three steps (\emph{concept discovery}, \emph{concept tagging}, and \emph{hierarchy building}) separately.

For each step, 500 data points are randomly sampled and divided into five groups with 100 data points each. On \emph{concept discovery}, a data point is an FoS; on \emph{concept tagging}, a data point is a concept-publication pair; and on \emph{hierarchy building}, a data point is a parent-child pair between two concepts. For the first two steps, each 100-data-points group is assigned to one human judge. 
The concept hierarchy results are by nature more controversial and prone to individual subjective bias, hence we assign each group of data to three judges and use majority voting to decide final results. 

The accuracy is calculated by counting positive labels in each 100-data-points group and averaging over 5 groups for each step. The overall accuracy is shown in Table \ref{tab:eval_accuracy_results} and some sampled hierarchical results are listed in Table \ref{tab:example_results}. 

Most hierarchy dissatisfaction is due to the intransitiveness and type-inconsistent limitations of the subsumption model. For example, most publications that discuss the \emph{polycystic kidney disease} also mention \emph{kidney}; however, for all publications that mentioned \emph{kidney}, only a small subset would mention \emph{polycystic kidney disease}. According to the subsumption model, \emph{polycystic kidney disease} is the child of \emph{kidney}. It is not legitimate for a \emph{disease} as the child of an \emph{organ}. Leveraging the entity type information to fine-tune hierarchy results is in our plan to improve the quality.

\section{Conclusion}

In this work, we demonstrated a Web-scale production system that enables an easy exploration of scientific knowledge. 
We designed a system with three modules: concept discovery, concept tagging to  publications, and concept hierarchy construction. 
The system is able to cover latest scientific knowledge from the Web and allows fast iterations on new algorithms for document and language understanding. 

The system shown in this paper builds the largest cross-domain scientific concept ontology published to date, and it is one of the core components in the construction of the Microsoft Academic Graph, which is a publicly available academic knowledge graph---a data asset with tremendous value that can be used for many tasks in domains like data mining, natural language understanding, science of science, and network science. 

\bibliography{acl2018}
\bibliographystyle{acl_natbib}

\end{document}